%% file: main.tex
\documentclass[lettersize,journal]{IEEEtran}
\usepackage{amsmath,amsfonts}
\usepackage{algorithmic}
\usepackage{algorithm}
\usepackage{array}
\usepackage[caption=false,font=normalsize,labelfont=sf,textfont=sf]{subfig}
\usepackage{textcomp}
\usepackage{stfloats}
\usepackage{url}
\usepackage{verbatim}
\usepackage{graphicx}
\usepackage{cite}

\usepackage{siunitx}
\usepackage{hyperref}
\usepackage{multirow}

\begin{document}

\title{
Multi-Modal Sensor Fusion and Object Tracking\\for Autonomous Racing
}

\author{
Phillip Karle, Felix Fent, Sebastian Huch, Florian Sauerbeck, and Markus Lienkamp
\thanks{
Manuscript received xx xx, 2023; revised xx xx, 2023; accepted
xx xx, 2023. Date of publication xx xx, 2023; date of current version xx xx, 2023.
This work was supported by the Bavarian Research Foundation, the Munich Cluster for the Future of Mobility, and MAN Truck \& Bus SE.

Corresponding author: Phillip Karle.

The authors are with the Institute of Automotive Technology, Technical University of Munich,
85748 Garching, Germany (e-mail: phillip.karle@tum.de, felix.fent@tum.de, sebastian.huch@tum.de, florian.sauerbeck@tum.de).

Digital Object Identifier ...
}%
}

\markboth{IEEE TRANSACTIONS ON INTELLIGENT VEHICLES,~Vol.~14, No.~8, August~2021}%
{Shell \MakeLowercase{\textit{et al.}}: A Sample Article Using IEEEtran.cls for IEEE Journals}

\IEEEpubid{0000--0000/00\$00.00~\copyright~2021 IEEE}

\maketitle

\input{source/00_abstract}

\input{source/01_introduction}

\input{source/02_related_work}

\input{source/03_method}

\input{source/04_results}

\input{source/05_conclusion_contrib}

\newpage
\input{source/98_appendix}

\bibliographystyle{IEEEtran}
\bibliography{references}

\input{source/99_biography}

\end{document}

%% file: source/00_abstract.tex
\begin{abstract}
Reliable detection and tracking of surrounding objects are indispensable for comprehensive motion prediction and planning of autonomous vehicles.
Due to the limitations of individual sensors, the fusion of multiple sensor modalities is required to improve the overall detection capabilities.
Additionally, robust motion tracking is essential for reducing the effect of sensor noise and improving state estimation accuracy. 
The reliability of the autonomous vehicle software becomes even more relevant in complex, adversarial high-speed scenarios at the vehicle handling limits in autonomous racing.
In this paper, we present a modular multi-modal sensor fusion and tracking method for high-speed applications.
The method is based on the Extended Kalman Filter (EKF) and is capable of fusing heterogeneous detection inputs to track surrounding objects consistently. A novel delay compensation approach enables to reduce the influence of the perception software latency and to output an updated object list.
It is the first fusion and tracking method validated in high-speed real-world scenarios at the Indy Autonomous Challenge 2021 and the Autonomous Challenge at CES (AC@CES) 2022, proving its robustness and computational efficiency on embedded systems.
It does not require any labeled data and achieves position tracking residuals below 0.1\,m. The related code is available as open-source software at \url{https://github.com/TUMFTM/FusionTracking}.
\end{abstract}

\begin{IEEEkeywords}
Sensor Fusion, Multi-Object Tracking (MOT), Data Association, Extended Kalman Filter, Autonomous Vehicles
\end{IEEEkeywords}

%% file: source/01_introduction.tex
\section{Introduction}
Autonomous racing is one of the great accelerators of autonomous driving.
The dedicated application of autonomy software on the race track offers a challenging real-world evaluation.
Novel methods can be quickly validated in a full software stack before being transferred to public roads.
A robust method to reliably detect and track surrounding objects at high speeds and with low overall software latency is one of the open research questions in this field \cite{Betz2022}.
Similar to autonomous driving on public roads, each type of prevalent sensor has its limitations if being used alone.
This indicates that fusing multiple sensor modalities is necessary \cite{Wang.2020}.
Besides that, consistent tracking of surrounding objects is essential to enable accurate and reliable trajectory prediction and ego-motion planning \cite{Leon2021}.
\IEEEpubidadjcol
Lastly, the compensation of perception software delay given the constraint of limited hardware resources is of high relevance, especially at high speeds with low reaction times.

Based on these issues, we formulate our problem as follows: We want to contribute a robust fusion and tracking method, which reliably handles multiple heterogeneous sensor modalities and consistently and accurately tracks the motion of surrounding objects. The method shall be real-world applicable in a software stack for autonomous driving at high speeds, which requires low latency and the consideration of perception software delay. In addition, it shall be applicable without the need for labeled data. Our late-fusion and object tracking method to solve this problem formulation is built up as follows.
The multi-modal late fusion can handle input from multiple, heterogeneous detection pipelines. The raw input is filtered for out-of-track objects and multiple detections per object.
Then distance-based matching associates the filtered object lists with the currently tracked objects in chronological order. 
If a successful match occurs, the Extended Kalman Filter (EKF) is applied to a kinematic motion model for state estimation.
An important feature of the method is the delay compensation: Due to the delay of the detection input, a backward search in the observation storage is applied to get the tracked objects at the sensor timestamp. The optimized state estimation of the historic state is then forward integrated with the kinematic model to update all storage entries up to the current timestamp. By this, motion prediction and ego-motion planning receive updated, optimized tracks of the surrounding objects.
\begin{figure}
    \centering
    \includegraphics[clip, trim=0.2cm 0.5cm 0.2cm 0.2cm, width=1.0\columnwidth]{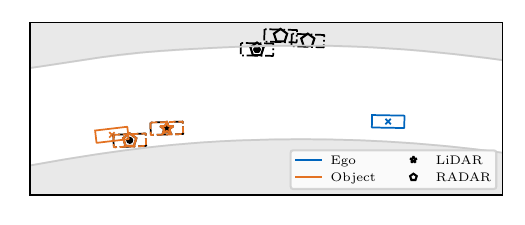}
    \caption{Real-world scenario of the proposed multi-modal object fusion and tracking approach at the AC@CES 2022 (driving direction: Left).}
    \label{fig:concept}
\end{figure}
Figure~\ref{fig:concept} shows a typical scenario of the proposed approach recorded at the Autonomous Challenge at CES (AC@CES) 2022 including the out-of-track-filter (black), the delayed perception (orange dashed), and the compensated optimized state estimation (orange solid).
To conclude, the main contributions of this work are:
\begin{itemize}
    \item A modular late fusion method for heterogeneous detection modalities.
    \item A perception software delay compensation by kinematic backward-forward integration.
    \item A tracking algorithm validated in real-world application in a full AV software for autonomous racing \cite{betz2022tum, wischnewski2022indy} at speeds up to \SI{270}{\kilo\metre\per\hour}.
\end{itemize}

%% file: source/02_related_work.tex
\section{Related Work} \label{sec:relatedwork}
This section provides a literature review of the state of the art in sensor fusion (\ref{sec:latefusion}) and multi-object tracking (MOT) (\ref{sec:mot}), and then a concluding evaluation of the review is given.

\subsection{Sensor Fusion} \label{sec:latefusion}
Sensor fusion describes the process of merging data from heterogeneous sensor modalities to improve particular criteria for decision tasks \cite{fayyad2020deep}. The three common approaches for sensor fusion are early fusion at raw-data level, mid-level fusion at feature level and late fusion at decision level \cite{Kolar2020}.

Wang et al. \cite{Wang.2020} analyze the limitations of common sensor modalities. The authors state that RADAR lacks a high 3D resolution of the environment, while the camera and LiDAR have disadvantages in poor weather conditions. They conclude that the fusion of heterogeneous sensor data is necessary to overcome the respective shortcomings.
As stated by Malawade et al. \cite{malawade2022hydrafusion}, early fusion enables the extraction of multi-modal features, but it is more sensitive to noise and sensor outliers, which reduces the robustness of this method. In contrast, late fusion approaches are more robust, but the mean Average Precision (mAP) is limited because intermediate features cannot be combined across the sensors.
De Jong et al. \cite{deJong2021} confirm this finding and conclude that the classification performance of late fusion approaches is lower compared to early fusion. However, they emphasize the reduced complexity of late fusion in comparison with early and mid-level fusion methods.

Early fusion approaches are mainly realized by the fusion of LiDAR and camera. In the recent year, mainly deep learning methods were used for this kind of early fusion \cite{xiao2015crf, xiao2018hybrid, sobh2018end, Xu2018PointFusionDS, Qi2018, Liang_2019_CVPR, Sindagi2019, Ku2018, Liang_2018_ECCV}. Notable approaches of 2D RADAR data fusion methods are given in \cite{Chadwick.2019, Nobis.2019, Kowol.07.10.2020}.
The fusion is performed with the camera modality at a low data level using deep learning methods, but the accuracy of these fusion networks is lower than the accuracy of LiDAR-camera methods \cite{Nobis.2021}.
A mid-level fusion approach with a combination of two modalities on various stages is presented by Nie et al. \cite{Nie2021}. The authors present a mid-level fusion and motion prediction approach. Separate networks are trained for the respective perception modality. Their features are fused by a dedicated fusion network and various stages. The approach is validated on the KITTI data set \cite{geiger2012we} and shows superior performance compared to the state of the art on this data set.

In the following, the review of the state of the art is focused on late fusion methods. For further information and reviews of sensor fusion techniques, the reader is referred to \cite{ Leon2021, Jimenezbravo.2022, OUNOUGHI2023267}.
An analysis of the application of multi-sensor fusion approaches on an embedded platform is given in \cite{jagannathan2018multi}.

A comparison of different filter techniques for a late fusion approach with a joint probabilistic data association of a 2D LiDAR and camera perception system is proposed by Garcia et al. \cite{garcia2017sensor}. The results of the conducted real-world tests reveal that the Unscented Kalman Filter (UKF) \cite{wan2001unscented} and Particle Filter (PF) \cite{del1997nonlinear} perform better than the Kalman Filter (KF) \cite{kalman1960new} with computational advantages accruing to the UKF compared to the PF.
Jahromi et al. \cite{Jahromi2019} demonstrate a real-world application of a late fusion approach by fusing independent detection pipelines using an EKF \cite{sorenson1985, jazwinski2007stochastic}. The authors report an improvement in the root mean square error (RMSE) compared to single-sensor pipelines. However, the real-world application is a simple low-speed maneuver and the calculation time is at \SI{200}{\ms}.
In comparison, Verma et al. \cite{Verma2018} present a real-time detection and tracking system for an urban environment with LiDAR and mono-camera fusion. The deep-learning camera detection is enhanced by the fusion of LiDAR information. A constant velocity (CV) model is used to update the predicted state with the incoming measurements. The approach is evaluated in three real-world scenarios, including lane change, stop-and-go at intersections, and an overtaking maneuver. The authors report consistent tracking of the target object in the given test scenarios. However, the experiments are at low speed and with low ego-object distance and do not comprise transient behavior of detecting new and discarding old objects.
Granström et al. \cite{Granström2018} propose a stochastic sampling algorithm for data association. The method is validated in pedestrian tracking based on LiDAR data. The optimization of the implementation is mentioned as an outlook.
Kunjumon et al. \cite{Kunjumon2021} show a late-fusion approach of camera and LiDAR on the KITTI data set \cite{geiger2012we}. A Bayes fusion is used to optimize the position estimation of camera and LiDAR before the KF is applied. The results reveal that the optimization of the position estimation in combination with the Kalman filter shows better accuracy than the non-optimized fusion of sensor data.
Farag \cite{farag2021kalman} proposes a real-time object detection and tracking algorithm. LiDAR and RADAR data are processed by clustering algorithms and the resulting detections form the input of a kinematic state estimation model. The tracking performance of the UKF and the EKF are compared using real-world data with a lower tracking error, but higher complexity in parameter tuning and increased computation time  of the UKF.
Andert et al. \cite{andert2022accurate} investigate a method to adaptively adjust measurement uncertainty by a key predictor term, which outputs the measurement error as a function of measurement distance and angular position. The late fusion approaches of Garcia et al. \cite{garcia2017sensor} and Farag \cite{farag2021kalman} are used for the evaluation of the error prediction model. An improvement of \qty{42}{\percent} is reported on the conducted 1/10 scale autonomous vehicle (AV) tests with LiDAR and camera sensors. 
Muresan et al. \cite{Muresan2020} present methods to stabilize and validate 3D object positions. Among them is a data association approach by comparing different motion models against each other and a fusion approach of a UKF with a neural network. The approaches are validated on real-world traffic. However, the recorded average position error is not acceptable for our use case. Besides that, no delay compensation is mentioned.

\subsection{Multi Object Tracking (MOT)} \label{sec:mot}
MOT is defined as the association of multiple detections of objects in a sequence of frames to estimate the objects' states \cite{Jimenezbravo.2022}. There are also approaches to combine the detection and tracking tasks \cite{ChavezGarcia2016}.
Rangesh and Trivedi \cite{Rangesh2019} present a multi-modal object tracking framework, which fuses cameras with varying field of views and LiDAR sensors. The framework extends an existing 2D MOT approach to 3D application.
Weng et al. \cite{Weng.2020} propose a real-time 3D MOT using a KF and the Hungarian algorithm \cite{Kuhn.1955} with the Intersection over Union (IoU) as metric. They achieve state-of-the-art results on the KITTI \cite{geiger2012we} and nuScenes benchmark data sets \cite{caesar2020nuscenes}. However, the authors use only LiDAR input data. No multi-modal data are given.
Chiu et al. \cite{Chiu.16.01.2020, chiu2021probabilistic} combine a KF-based state estimation and a data association by use of the Hungarian Method with the Mahalanobis distance \cite{DEMAESSCHALCK20001} as cost function. They achieve an Average Multi-Object Tracking Accuracy (AMOTA) of 0.55 on the NuScenes data set.
The association through Intersection of Union (IoU) as cost function is proposed by Weng et al. \cite{Weng.2020}. The authors report a worse AMOTA of 0.454 on the NuScenes data set. However, they note that the method reduces the calculation time and consequently is more suitable for real-world applications.
The two-stage data association and tracking method of Dao et al. \cite{Dao.2021} for image-based 3D tracking achieves an AMOTA of 0.587 on the NuScenes data set \cite{caesar2020nuscenes} and an AMOTA of 0.365 on the Waymo tracking data set \cite{sun2020scalability}.
Guo and Zhao \cite{Guo2023} propose an adaptive cubature Kalman filter to remove the influence of unknown bias. In addition, the authors introduce a new affinity model for data association, which evaluates the similarity between trajectories and candidate detections as a bipartite matching problem. The results show improvements in accuracy and reliability. However, the algorithm makes only use of LiDAR.

In addition to methods based on Bayesian filtering, deep learning-based MOT has recently emerged. The methods can be classified into deep network features-based MOT enhancement, deep network embedding, and deep network learning. For further information on these methods, the reader is referred to \cite{Pal.2021, WenhanLuo.2021}. 

In summary, although deep learning approaches achieve high tracking accuracy on dedicated data sets, the algorithms are limited to the specific operational design domain (ODD) the data set originates from, a fixed sensor setup, require labeled data, and have high training and calculation effort. Therefore, a robust real-world application is questionable \cite{shafiee2020deep}. Besides that, the accuracy is strongly dependent on the quality of the data set including the labeling, which requires time and high costs. The same is true for integrative deep learning early fusion techniques. Additionally, the black box character of deep learning-based algorithms also implies limited interpretability and robustness.

\begin{figure*}
    \centering
    \includegraphics[width=1.99\columnwidth]{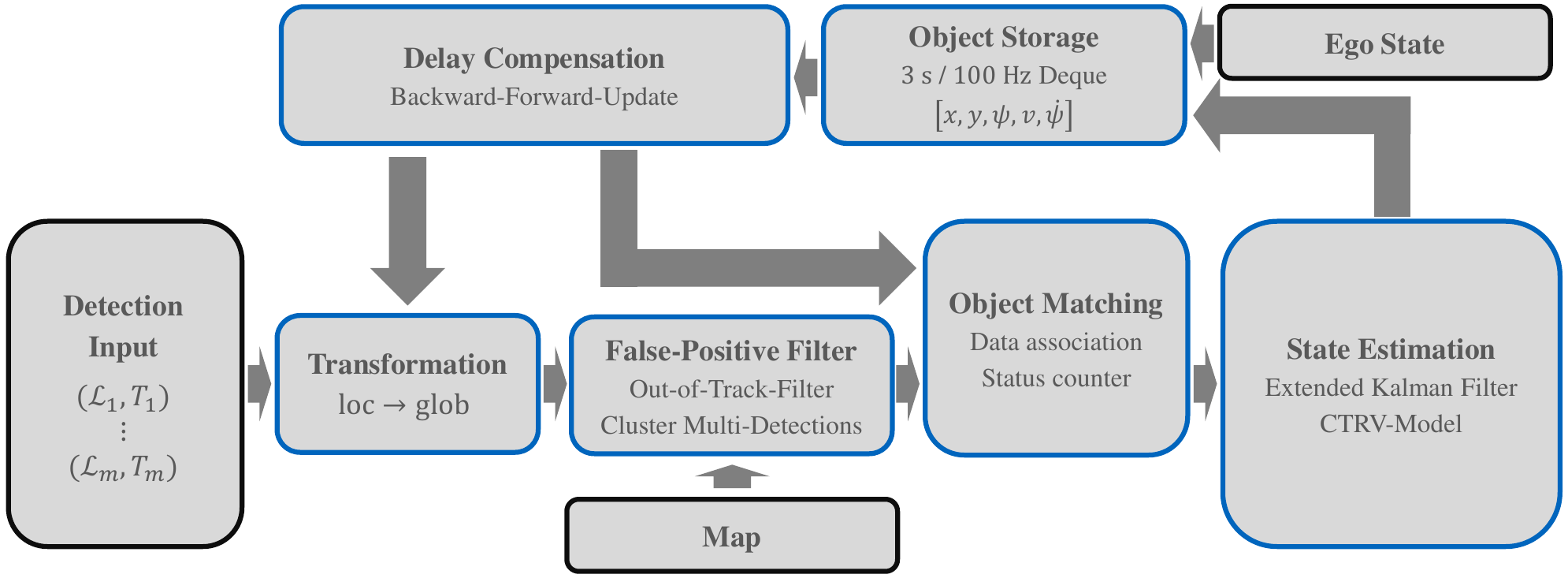}
    \caption{Interfaces (black) and structure of the multi-modal sensor fusion and object tracking method.}
    \label{structure}
\end{figure*}

Furthermore, the reported state-of-the-art real-world applications of fusion and tracking methods are at low speeds and in simple scenarios. As stated by Betz et al. \cite{Betz2022}, high-speed applications of fusion techniques are currently lacking in the field of autonomous driving. The authors identify high-speed perception as one of the major challenges in autonomous racing and list a reduction in computational delay and an improvement of sensor fusion performance as key challenges.
The overall objective of enabling reliable high-speed driving is directly related to the tolerable latency of the system as stated by Falange et al. \cite{Falanga2019}, who analyze the dependency of maximal latency and safety guarantees in a case study of autonomous quadrocopters. Even though there is no analysis specifically conducted for autonomous racing, the constraints for both types of autonomous system applications are comparable, because of the highly dynamic environment.

Lastly, there is no method in the presented state of the art that explicitly compensates for the perception delay.

Our work targets these aspects. We propose a method that is independent of any labeled data in development and application, can fuse heterogeneous detection input, and compensate for perception delay to reliably track surrounding objects.

%% file: source/03_method.tex
\section{Methodology} \label{sec:multimodelobjectfusion}
The following section describes the proposed method of the multi-modal late fusion and tracking approach.
It comprises the processing of the detection input (\ref{sec:input}), the transformation (\ref{sec:trafo}), and the False-Positive filters (\ref{sec:plausibilitycheck}), followed by the object matching step (\ref{sec:matching}). Subsequently, the state estimation is conducted (\ref{sec:stateestimation}) before the optimized state is stored (\ref{sec:storage}). Additionally, the delay compensation is described, which reduces the effect of the perception latency to output an updated object list to the succeeding modules (\ref{sec:delaycompensation}).
Fig.~\ref{structure} gives a schematic diagram of the proposed method.
Furthermore, the data used for the evaluation are investigated in this section (\ref{sec:data}).
For all parameters of the described method, the reader is referred to Table \ref{tab:parameter_tracking} in the appendix.

\subsection{Input Processing} \label{sec:input}
The inputs to the proposed method are the dynamic state of the ego vehicle, a map, which contains the track boundaries in vector form, and a variable number of object lists from the respective detection pipelines $m$. The detection pipelines are flexible and can be added modularly. There is no limitation to a specific sensor modality. In addition to the detected objects, specified by the measured features in the object list $\mathcal{L}_i$, the sensor detection time stamp $T_i$ is also input, which is highly relevant for the data association step. The multi-modality of the approach allows different measured features and varying cycle times for each sensor modality. By this, heterogeneous sensor modalities can be combined and their individual drawbacks can be compensated.

The detection input tuples $\left( \mathcal{L}_i, T_i \right)$ are processed chronologically from the oldest to the newest detection time stamp. Each detection pipeline is configured by its measurement features, related uncertainties, and a weighted match counter. This counter is added to the status counter of the respective object if a successful match occurs. The status counter represents the number of detection inputs without a successful match until the object is removed from storage. The individual weight of the match counter per detection modality makes it possible to consider the specificity of each modality.

\subsection{Transformation} \label{sec:trafo}
After the detection input is received, a transformation from the local (vehicle-based) to the global coordinate system takes place.
To reduce the influence of high-frequency sensor noise in the estimated ego state, a low-pass filter is applied to the received ego state before it is used for the coordinate transformation.

\subsection{False-Positive Filter} \label{sec:plausibilitycheck}
\begin{figure*}
	\includegraphics[clip, trim=0.0cm 0.3cm 0.0cm 0.1cm, width=1.99\columnwidth]{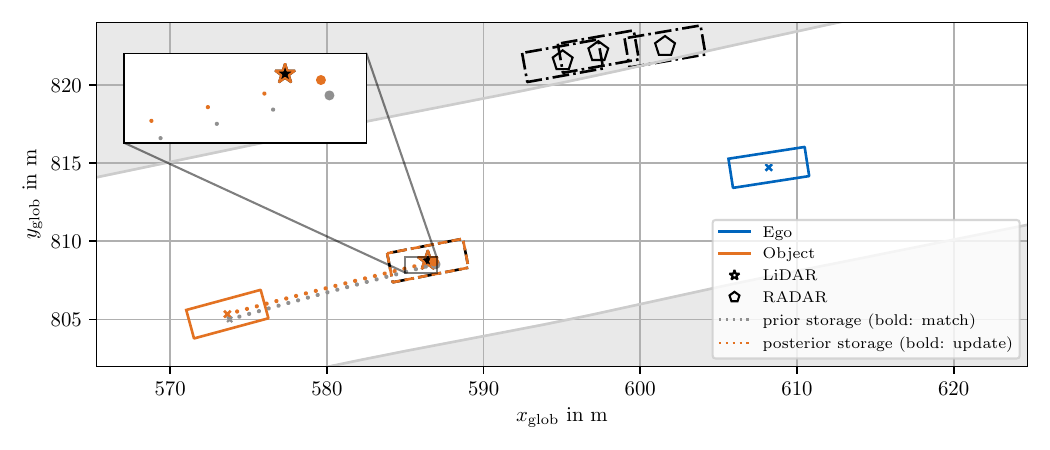}
	\caption{Exemplary high-speed real-world scenario at AC@CES 2022. The ego speed is \SI{255}{\kilo\metre\per\hour}, the object speed is \SI{233}{\kilo\metre\per\hour}.}
	\label{fig:delay_comp}
\end{figure*}

The transformed object lists are filtered using two plausibility checks to reduce the number of FPs. First, the received map (containing the driveable area) is used to remove all objects outside the track. This assumption holds true for the given use case on a closed race track. Fig. \ref{fig:delay_comp} shows an exemplary scenario with out-of-track filtered objects (black).

Besides that, a plausibility check is conducted within the object list to check if single detections of one modality overlap each other. For this purpose, a k-d tree clustering \cite{Maneewongvatana1999} is applied with a fixed distance threshold.

\subsection{Object Matching} \label{sec:matching}
Object matching associates the newly detected objects with the old predicted objects, which are currently tracked. Due to the given high-speed application and an expected detection delay, the Euclidean distance is applied as a cost function to match the objects in a pairwise manner. The combinatorial task to obtain the overall minimum distance costs between the $n$ old tracked and $m$ new detected objects is solved using the Hungarian method \cite{Kuhn.1955}. The solution to the assignment problem can result in the following three cases:

\begin{enumerate}
    \item \textbf{New Unmatched Object} \\
    Either the detection object list contains more elements than currently tracked objects ($n < m$) or the inter-object distance exceeds the maximum matching distance, i.e. no valid match results. In these cases, a new object is initialized. The initialization comprises the creation of a unique object ID, the setup of the status counter, and the initialization of the object's individual state estimation.
    
    \item \textbf{Old Matched Object} \\
    The inter-object distance is below the maximum matching distance, i.e., a successful match is given. In this case, the update step of the state estimation is conducted and the status counter is increased. The corrected posterior state is written to object storage.
    
    \item \textbf{Old Unmatched Object} \\
    Either there are more currently tracked objects ($n > m$) than the detection object list contains or the inter-object distance is above the maximal matching distance. In these cases, the status counter of the object is decreased and the predicted object state is written to the object storage without a measurement update. If the status counter of the object is zero, the object is removed from the object storage.
\end{enumerate}
The status counter referred to, is designed to weigh the sensitivity and specificity of the detection pipelines against each other. The matching performance during the transition to reliably track a new object and to remove an object that is not detected anymore is decisively influenced by this counter. The counter is a positive integer and indicates how many times an unsuccessful match (case 3) can occur before the object is discarded, i.e. the counter is decreased by 1 if an old object is not matched. In the event of a successful match (case 2), the detection pipeline-specific match counter value is added to the status counter. The parameterized upper threshold of the status counter ensures that the maximum number of unsuccessful matches before object removal is limited. It is important to mention that there is no probabilistic association, hence the fixed distance threshold and the status counter are strictly applied. 

\subsection{State Estimation} \label{sec:stateestimation}
The state estimation of the tracked object is realized by an EKF. It is chosen because of its ability to approximate non-linear models with acceptable accuracy compared to the linear KF and its low computational cost and interpretable parameter optimization, which is possible without the availability of rich data compared to the UKF \cite{laviola2003comparison}.
The EKF expects a state-space model:
\begin{equation}
\boldsymbol{x}_{k} = f(\boldsymbol{x}_{k-1}, \boldsymbol{u}_{k}) + \boldsymbol{w}_{k}
\end{equation}
\begin{equation}
\boldsymbol{z}_{k} = h(\boldsymbol{x}_{k}) + \boldsymbol{v}_{k}
\end{equation}
where $\boldsymbol{x}_{k}$ is the state vector at step $k$, which is determined by the state transition model $f(\boldsymbol{x}_{k-1}, \boldsymbol{u}_{k})$ based on the old state variables and the control vector $\boldsymbol{u}_{k}$. The observation vector $\boldsymbol{z}_{k}$ is calculated by the observation model $h(\boldsymbol{x}_{k})$. Process and observation noise ($\boldsymbol{w}_{k}$, $\boldsymbol{v}_{k}$) are assumed to be zero-mean multivariate Gaussian noise and are considered in the prediction and update step (eq. \ref{eq:pkk}, \ref{eq:sk}). The alternating process of prediction (prior) and update step (posterior) of the state estimation is as follows:

\textbf{Predict}
\begin{equation}
    \hat{\boldsymbol{x}}_{k|k-1} = f(\hat{\boldsymbol{x}}_{k-1|k-1}, \boldsymbol{u}_{k}) \label{eq:xkk}
\end{equation}
\begin{equation}
    \boldsymbol{P}_{k|k-1} =  {{\boldsymbol{F}_{k}}} \boldsymbol{P}_{k-1|k-1}{ {\boldsymbol{F}_{k}^\top}} + \boldsymbol{Q}_{k}
    \label{eq:pkk}
\end{equation}

\textbf{Update}
\begin{equation}
    \tilde{\boldsymbol{y}}_{k} = \boldsymbol{z}_{k} - h(\hat{\boldsymbol{x}}_{k|k-1}) \label{eq:res}
\end{equation}
\begin{equation}
    \boldsymbol{S}_{k} = {{\boldsymbol{H}_{k}}}\boldsymbol{P}_{k|k-1}{{\boldsymbol{H}_{k}^\top}} + \boldsymbol{R}_{k} 
    \label{eq:sk}
\end{equation}
\begin{equation}
    \boldsymbol{K}_{k} = \boldsymbol{P}_{k|k-1}{{\boldsymbol{H}_{k}^\top}}\boldsymbol{S}_{k}^{-1}
\end{equation}
\begin{equation}
    \hat{\boldsymbol{x}}_{k|k} = \hat{\boldsymbol{x}}_{k|k-1} + \boldsymbol{K}_{k}\tilde{\boldsymbol{y}}_{k}
\end{equation}
\begin{equation}
     \boldsymbol{P}_{k|k} = (\boldsymbol{I} - \boldsymbol{K}_{k} {{\boldsymbol{H}_{k}}}) \boldsymbol{P}_{k|k-1}   \label{eq:p_kk}
\end{equation}
where $\hat{\boldsymbol{x}}_k$ denotes the estimation of the state vector $\boldsymbol{x}_k$ at timestep $k$, $\tilde{\boldsymbol{y}}_k$ is the measurement residual between the measurement $\boldsymbol{\Tilde{z}_k}$ and the measurement prediction based on the prior state $\boldsymbol{h}(\boldsymbol{\Tilde{x}_{k|k-1}})$. $\boldsymbol{R}_{k}$ and $\boldsymbol{Q}_{k}$ represent the covariance matrices of the process and observation noise. The state transition and the observation matrices ($\boldsymbol{F}_{k}$, $\boldsymbol{H}_{k}$) are the first-order partial derivatives of the state transition and the observation model:
\begin{equation}
    {{\boldsymbol{F}_{k}}} = \left . \frac{\partial f}{\partial \boldsymbol{x} } \right \vert _{\hat{\boldsymbol{x}}_{k-1|k-1},\boldsymbol{u}_{k}}
\end{equation}
\begin{equation}
    {{\boldsymbol{H}_{k}}} = \left . \frac{\partial h}{\partial \boldsymbol{x} } \right \vert _{\hat{\boldsymbol{x}}_{k|k-1}} 
\end{equation}

The state-space model used is a kinematic, point-mass model, which assumes the constant turn rate and velocity (CTRV-model) of the object. The model is chosen to incorporate the yaw rate so that it can model turn movements accurately. The acceleration as a further common state variable is neglected because it is not measured and by this, the estimation is prone to oscillations. The model equations are as follows:
\begin{equation}
    \begin{pmatrix}
    x\\
    y\\
    \psi\\
    v\\
    \dot{\psi}
    \end{pmatrix}_{k}
    =
    \begin{pmatrix}
    x -v \Delta t \sin{\psi}\\
    y + v \Delta t \cos{\psi}\\
    \psi + \Delta t \dot{\psi}\\
    v\\
    \dot{\psi}
    \end{pmatrix}_{k-1}
    \label{eq:ctrv}
\end{equation}
By default, the timestep is set to $\Delta t = 1 / f_{\mathrm{EKF}}$ with  $f_{\mathrm{EKF}} = \SI{100}{\hertz}$. The small step size reduces the approximation error of the model assumption, especially if objects are cornering because the yaw angle is updated by the yaw rate estimate for each timestep. It should be noted that the filter timestep is independent of the callback frequency of the overall module and the frequency of the detection pipelines. Thus, the prediction step of the state estimation is constantly called to iteratively estimate the state of the tracked objects. In this way, the objects' states are equidistantly sampled and stored in the object storage. The update step of the filter is only called if a successful match (case 2) exists. This procedure enhances the modularity of the module so that it can cope with irregular or missing detection inputs.

The initialization of the estimated object state is of high importance in reducing the transient phase if an object enters the detection range. While the measured features of the respective detection pipelines are directly used for the state initialization, further assumptions are made to improve the estimation of the initial state. The first assumption is that all detected vehicles drive in the same direction, which is valid for the race track application. Thus, the orientation of the track's centerline is used as an initial guess for the object's yaw angle. The second assumption is that the object's speed is of the same magnitude as the ego speed. So the ego speed is multiplied with the scalar value $k_v$ to guess the object's speed. 
The coarse estimation of these two state variables is applied with high measurement uncertainty to adjust them in a few EKF update steps based on the kinematic dependencies. It should be noted that the yaw angle assumption is also applied for every input from a detection pipeline without detected yaw rate, e.g. the RADAR.

\subsection{Object storage} \label{sec:storage}
The object storage is a double-ended queue of fixed length. The length is determined by the EKF update frequency and the desired temporal object history. The default setting is a \SI{3}{\second} object history, which implies a storage length of 300 (for $f_{\mathrm{EKF}} =$ \SI{100}{\hertz}).

\subsection{Delay Compensation} \label{sec:delaycompensation}
In addition to the general effort to reduce the calculation time of a software module for a time-critical application such as autonomous driving, the proposed method incorporates a delay compensation functionality. This function aims to reduce the accumulated delay from sensor input up to the receipt of the extracted object lists by the presented state estimation. The so-called backward-forward integration works as follows.

The compensation function is called individually for the detection inputs from each sensor modality, which are sorted in chronological order starting with the oldest. The received sensor time stamp is used in the backward step to search the observation storage for all object states at the sensor time stamp. These past object states are used to match the given detection input.

In the event of a successful match, the update step of the EKF (eq. \ref{eq:res} - eq. \ref{eq:p_kk}) is applied to the historic object states. The optimized past object state is then used to update all subsequent states. This is realized by means of the forward integration of the kinematic model (eq. \ref{eq:xkk} - \ref{eq:pkk}). The forward integration is either done up to the time stamp of the next detection input or, if only one detection input is given, up to the present tracking time stamp. In the first case, the object matching function is called again to associate the next detection input with the updated object storage entries. Then the forward integration is continued with the corrected object states. By this backward-forward integration, it is possible to output an updated, high-frequency sampled tracked state, but still consider delayed detection inputs.

Fig. \ref{fig:delay_comp} shows an exemplary scenario. Before the received detection input is processed, the tracked object state is synchronized with the ego timestamp (grey cross) by forward integration (eq. \ref{eq:xkk} - \ref{eq:pkk}).
The only detection within the track boundaries is received from the LiDAR (orange-black star \& box) with a delay of \SI{211}{\ms}, i.e. at \SI{233}{\km\per\hour} the object has moved \SI{13.7}{\metre} since it was detected.
The implemented delay compensation considers the sensor delay and executes a backward search in the prior object storage (grey dotted).
The object storage entry with the smallest temporal difference to the detection timestamp is used for the matching (grey bold dot). In the given case, the difference is \SI{1.1}{\ms}. The inaccuracy due to the equidistant sampling is negligible, due to the high sampling frequency of \SI{100}{\hertz} (\ref{sec:storage}).
The past object state matches with the LiDAR detection (case 2), so it is the prior for the filter update. By means of the LiDAR detection, the measurement update is conducted. Then, the optimized object is forward integrated through the kinematic model to update all storage entries of the object (orange dotted).

\subsection{Evaluation Data} \label{sec:data}
\begin{figure*}
	\includegraphics [clip, trim=0.0cm 0.3cm 0.0cm 0.1cm, width=1.99\columnwidth]{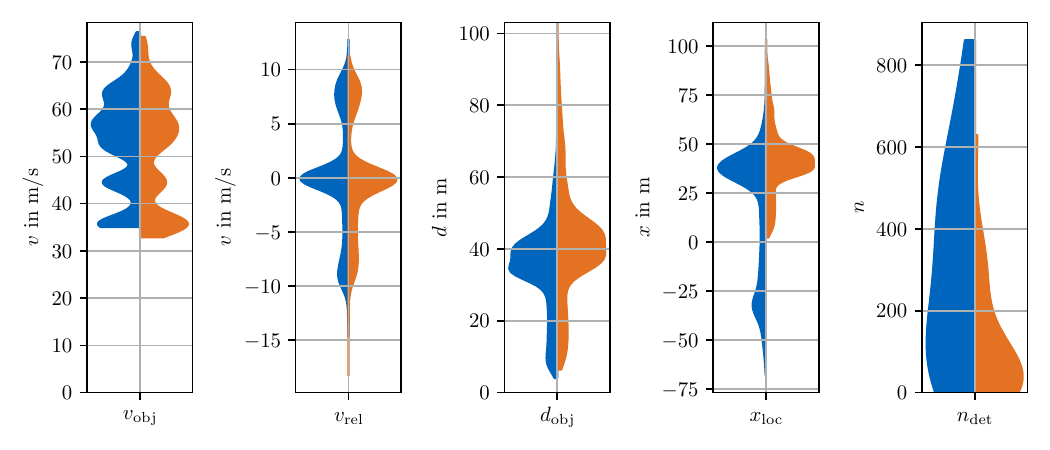}
	\caption{Data analysis of the input data from the LiDAR (blue) and RADAR (orange) detection.}
	\label{fig:data_analysis}
\end{figure*}

The data used for the evaluation of the proposed method are derived from the semifinal and final run of the TUM Autonomous Motorsport team at the AC@CES 2022 at the Las Vegas Motor Speedway (LVMS). The LVMS is an oval race track with banking between \SI{9}{\degree} and \SI{22}{\degree}. The mode of the AC@CES 2022 was a head-to-head overtaking competition of two opposing cars 
\cite{ACCESRules2022}. While the leading car had maintained a fixed target speed, the trailing car had to overtake within one lap. After a successful overtake, the roles were switched and the target speed of the leading car was increased.

The recorded data contain the fastest overtaking maneuvers of autonomous vehicles at the time of publication. The object lists provided by the LiDAR clustering (C) algorithm and the RADAR detection (R) algorithm are recorded in rosbags.

The data are available publicly, the respective links are given in the open-source repository. 
It should be noted that the data are unlabeled. So, only the objects detected by the perception algorithms are given and no ground truth states are available. However, it can be used as benchmark for delay compensation (ego timestamp available), consistency of object tracking at high speeds (total number of real objects known) and the robustness of the object state estimation (raw detections for comparison given).

In total, 11,052 samples of the LiDAR algorithm and 4,723 samples of the RADAR algorithm are recorded. From these samples, 28 valid objects are detected by the LiDAR and 26 valid objects by the RADAR. Valid is defined as an object that has been measured at least 3 times. Figure \ref{fig:data_analysis} outlines the data distribution over various metrics.

The object's speed $v_{\mathrm{obj}}$ is between \SI{35}{\metre\per\second} and \SI{75}{\metre\per\second}, which is the fastest overtaking speed in the data set. The distribution reflects the different speed stages the teams had to pass. The distribution of the relative speed between the ego and the detected object $v_{\mathrm{rel}}$ has a peak at \SI{0}{\metre\per\second}, which represents the situation in which one or other of the cars is driving behind the other. Positive and negative values relate to the speed difference when overtaking or being overtaken by the competing car.

The object's distance $d_{\mathrm{obj}}$ ranges up to \SI{105}{\metre} as the maximum detection range for the RADAR, and up to \SI{98}{\metre} for LiDAR detection. Most detections are within a distance of \SI{30}{\metre} to \SI{50}{\metre}. The longitudinal position related to the ego vehicle $x_{\mathrm{loc}}$ gives more insights about the relative position. The LiDAR algorithm detects an object up to \SI{57}{\metre} behind the car. The RADAR detections are only in front of the car because only the front RADAR was used in the data set.

The last violin plot shows the number of detections $n_{\mathrm{det}}$ per object and ranges from 3 measurements of non-consistently tracked objects up to 800 successfully matched detections.

%% file: source/04_results.tex
\section{Results} \label{sec:evaluation}
In the following section, the proposed method for multi-modal late fusion and object tracking is evaluated.
Initially, a comprehensive investigation of the different parts of the method is conducted, which comprises the evaluation of the overall tracking performance of the EKF (\ref{sec:tracking_acc}), the ability to compensate for delay (\ref{sec:res_delay_compensation}) and the transient behavior of the module (\ref{sec:transient_behavior}).
The second part of the evaluation presents a variation of several parameters to analyze the sensitivity of the method (\ref{sec:sensitivity}).
The evaluation is conducted on high-speed real-world data for autonomous racing. An analysis of the data is given in \ref{sec:data}.

\subsection{Tracking Performance} \label{sec:tracking_acc}
The tracking performance of the EKF is evaluated by the measurement residual (eq. \ref{eq:res}) and the precision. 
The residual is an established method to evaluate state estimation methods and is chosen because no ground truth data is available.
In the case given, the residual comprises the longitudinal and lateral position, the yaw angle, and the velocity (RADAR detections only). Derived from the mathematical theory of the EKF, optimal filter tuning is achieved when the residuals are zero-mean Gaussian distributed with the same standard deviation as assumed in the observation noise \cite{powell2002automated}. In this case, the noisy measurement is successfully filtered and the expected value is obtained.

Additionally, the precision of the algorithm is analyzed, which is defined as:
\begin{equation}
    \mathrm{Precision} = \frac{\mathrm{TP}}{\mathrm{FP}+\mathrm{TP}}
\end{equation}
with the object counts of True Positives (TP) and False Positives (FP). The number of TP objects is determined manually for the given data source.
The TPs are all objects within the sensor range, i.e. detectable by the sensors. A TP object gets a unique identifier (UID) while it's inside the detectable area. Thus, in contrast to the framewise view of detection benchmarks, we consider the whole scenario to evaluate the precision of the algorithm.
The FPs are all ghost objects that are set up with an UID or real objects, which are only tracked for a short time and then discarded by the algorithm.
As a detailed analysis of the TPs and FPs reveals, the FPs are mainly objects that are tracked for a short time at the edge of the sensor range but are then lost due to unstable tracking without successful re-matches.

We want to emphasize that this definition of TPs and FPs differs from the framewise definition in object detection. However, our definition is more suitable for the given evaluation of the fusion and tracking algorithm because the consistency of the algorithm is considered. An inconsistent tracking of surrounding strongly degrades the overall software performance because stable overtaking maneuvers can not be realized.

Due to the design of the algorithm, for early tracking of competing cars, even if there are not yet in a stable detection range, an object is added to the observation storage after just two successful matches. This results in a high number of short-tracked objects, i.e., a high number of FPs and a low precision (Table \ref{tab:sensitivity}).

The tracking performance of our approach (EKF, CTRV) is compared to further kinematic models. As a baseline, a constant velocity (CV-model) and a constant turn-rate and acceleration (CTRA-model) are implemented. The consideration of linear model and the usage of the KF for comparison was discarded because of the high dependency on the heading estimation for high-speed state estimation.

As shown in Table \ref{tab:sensitivity} (row "EKF, CTRV"), the mean of the longitudinal residual is slightly negative, which results from the speed residual, which also has a negative mean. The standard deviation of the longitudinal residual of \SI{0.73}{\metre} also results from the high standard deviation of the objects' speed. However, compared to the ego-object distance (Fig. \ref{fig:data_analysis}), the deviation is low enough for stable tracking.

The lateral residual is close to a zero-mean and its standard deviation implies that the majority of the lateral residuals are within $\pm$\SI{1}{\metre} ($3 \sigma$-rule). The residual of the heading has a high standard deviation, which reflects the fact that the heading of the track is used, and neither the LiDAR nor the RADAR measures the object's heading. The residual of the object's velocity is entirely based on the RADAR measurement. Hence, the high standard deviation of \SI{2.76}{\metre\per\second} is directly related to the accuracy of the RADAR speed measurement.

Overall, the low-biased residuals of the observation features show that the parameterization is properly set for the given detection inputs. The low lateral residuals in particular are of high relevance for the given use case of wheel-to-wheel racing. However, the volatile behavior of the speed estimation can result in poor tracking accuracy in transient situations with high acceleration. The low precision of \SI{50}{\percent} indicates that multiple objects are initiated until a new object is tracked stably.

In comparison, the CV-model has a high bias of the lateral residual, resulting from the bias of the yaw residual. It shows that neglecting the turn rate has a high impact, as object motion in turns is not modeled sufficiently. The simplified motion model also results in a lower precision of \SI{-26}{\percent}. The accuracy of the CTRA-model is equal in lateral direction and better in longitudinal direction. Hence, even though the acceleration is not measured, the consideration in the state motion equations improves the longitudinal compensation. However, an evaluation of start scenarios revealed that the CTRA-model is prone to overshoot in longitudinal direction, especially if the speed is not measured, so we decided to use the CTRV-model for our application.

The average calculation time of the method on a single Intel i7-9850H CPU core is \SI{6.24}{\ms} with a \qty{90}{\percent}-quantile of \SI{9.27}{\ms}. For further details about the influence of the software latency on the overall software performance, the reader is referred to Betz et al. \cite{Betz2023}.

\subsection{Delay Compensation} \label{sec:res_delay_compensation}
\begin{figure}
    \centering
	\includegraphics[clip, trim=0.0cm 0.3cm 0.0cm 0.1cm, width=0.49\columnwidth]{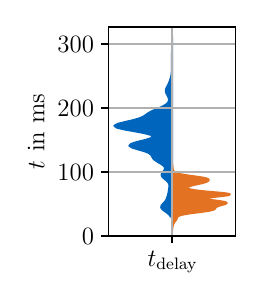}
	\includegraphics[clip, trim=0.0cm 0.3cm 0.0cm 0.1cm, width=0.49\columnwidth]{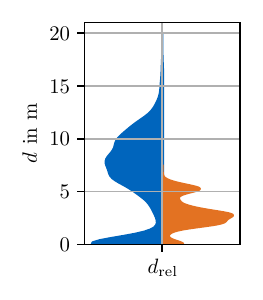}
	\caption{Distribution of delay from sensor time stamp to tracking subscription of LiDAR (blue) and RADAR (orange) in \SI{}{\ms} (left) and moved distance in \SI{}{\meter} (right).}
	\label{fig:avg_delay_comp}
\end{figure}

The evaluation of the compensated delay is given in Fig. \ref{fig:avg_delay_comp}. The average delay of the LiDAR is \SI{151}{\ms} with a \qty{90}{\percent}-quantile of \SI{191}{\ms}, the RADAR's average delay is \SI{62}{\ms}, the \qty{90}{\percent}-quantile is \SI{89}{\ms} (Fig. \ref{fig:avg_delay_comp}, left). Considering the ego speed, the moved distance of the ego object are in average \SI{6.20}{\metre} for the LiDAR and \SI{2.98}{\metre} for the RADAR sensor and ranges up to \SI{19.97}{\metre} (Fig. \ref{fig:avg_delay_comp}, right) in maximum. 

The related tracking performance with the delay-compensated detection inputs is implicitly given in the overall tracking performance, as the given real-world data comprises the original sensor time stamps. Hence, the low-biased residuals and solid tracking accuracy prove the efficacy of the delay compensation method (Table \ref{tab:sensitivity}).

\subsection{Transient behavior} \label{sec:transient_behavior}
\begin{figure*}[h]
	\includegraphics[clip, trim=0.0cm 0.3cm 0.0cm 0.1cm, width=1.99\columnwidth]{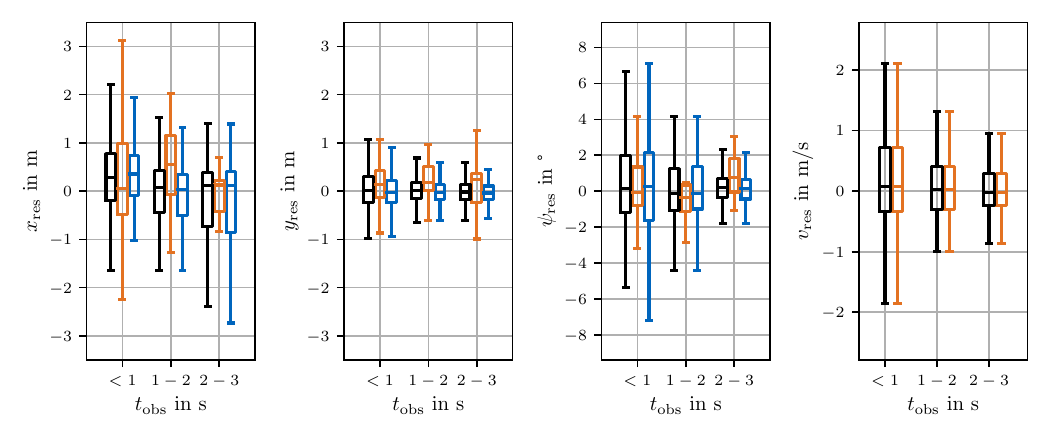}
	\caption{Residual error in different observation times of the fused system (black), of the LiDAR (blue) and RADAR (orange) detection inputs.}
	\label{fig:res_vs_t_obs}
\end{figure*}
In addition to the overall tracking performance, the transient behavior of the filter is of particular interest. To investigate the transient phase of the EKF, we evaluated the residuals separated by the time since the start of object tracking.
As Fig. \ref{fig:res_vs_t_obs} reveals, the residuals of the lateral position $y_{\mathrm{res}}$ and the yaw angle $\psi_{\mathrm{res}}$ decrease during the first \SI{3}{\second}. 
The same is true for the velocity residual of the RADAR measurements. However, the longitudinal position residuals, especially of the LiDAR clustering algorithm, exhibit a volatile behavior. Thus, in the given data, the longitudinal position estimate has not reached a steady state within the given time horizon. The high lateral residuals of the RADAR sensor are caused by the low tangential accuracy of this modality. Overall, the high residuals at observation times below \SI{1}{\second} indicate that the filter parameter could be further optimized, e.g. with adaptive values of process and measurement noise values.

\subsection{Sensitivity Analysis} \label{sec:sensitivity}

\newcolumntype{M}[1]{>{\centering\arraybackslash}m{#1}}
\begin{table*}[b!]
    \caption{Sensitivity analysis of the measurement residuals (mean, std) and precision.}
    \centering
    \begin{tabular}{|l|M{0.02\textwidth}|M{0.05\textwidth}|M{0.05\textwidth}|M{0.05\textwidth}|M{0.05\textwidth}|M{0.05\textwidth}|M{0.05\textwidth}|M{0.05\textwidth}|M{0.05\textwidth}|M{0.05\textwidth}|M{0.05\textwidth}|}
    \hline
    \multicolumn{2}{|c|}{} &
    \multicolumn{2}{|c|}{$x_{\mathrm{loc}}$ in \SI{}{\metre}} &
    \multicolumn{2}{c|}{$y_{\mathrm{loc}}$ in \SI{}{\metre}} &
    \multicolumn{2}{c|}{$\psi$ in \SI{}{\degree}} &
    \multicolumn{2}{c|}{$v$ in \SI{}{\metre\per\second}} &
    \multicolumn{2}{c|}{{Precision}}
    \\
	\cline{3-12}

    \multicolumn{2}{|c|}{} &
    $\mu$ &
    $\sigma$&
	$\mu$ &
	$\sigma$&
	$\mu$ &
	$\sigma$ &
	$\mu$ &
	$\sigma$ &
	abs. &
	rel.
	\\
	\hline\hline
	
    EKF, CTRV & $-$ & $-0.08$ & $0.73$ & $0.03$ & $0.38$ & $-0.11$ & $2.58$ & $-0.21$ & $2.76$ & $0.5$ & $-$ \\ \hline
    EKF, CV & $-$ & $ -0.10 $ & $ 0.72 $ & $ 0.63 $ & $ 0.56 $ & $ 3.15 $ & $ 2.89 $ & $ -0.13 $ & $ 2.94 $ & $ 0.37 $ & $ -0.26 $ \\ \hline
    EKF, CTRA & $-$ & $ -0.06 $ & $ 0.58 $ & $ 0.03 $ & $ 0.38 $ & $ -0.13 $ & $ 2.59 $ & $ -0.12 $ & $ 2.69 $ & $ 0.5 $ & $ 0.0 $ \\ \hline\hline
    
    \multirow{3}{*}{\shortstack[l]{Node frequency \\ $f_{\mathrm{node}}$ in \SI{}{\hertz}}}
    
    & 10 & $-0.03 $ & $ 0.80 $ & $ 0.02 $ & $ 0.35 $ & $ \boldsymbol{-0.07} $ & $ 2.69 $ & $ -0.24 $ & $ 3.30 $ & $ 0.42 $ & $ -0.15 $\\
    \cline{2-12}
    & 20 & $ -0.02 $ & $ 0.72 $ & $ 0.01 $ & $ 0.33 $ & $ -0.10 $ & $ 2.65 $ & $ -0.20 $ & $ 3.31 $ & $ 0.53 $ & $ 0.06 $\\
    \cline{2-12}
    & 100 & $-0.09 $ & $ 0.74 $ & $ 0.03 $ & $ 0.38 $ & $ -0.12 $ & $ 2.58 $ & $ -0.23 $ & $ 2.96 $ & $ 0.5 $ & $ 0.0 $\\

    \hline
    \hline

    \multirow{2}{*}{\shortstack[l]{Bounds filter \\ $d_{\mathrm{OBF, out}}$ in \SI{}{\metre}}}
    
    & 0 & $-0.08 $ & $ 0.74 $ & $ 0.03 $ & $ 0.40 $ & $ -0.12 $ & $\boldsymbol{2.55} $ & $ -0.23 $ & $ 3.38 $ & $ 0.42 $ & $ -0.15 $\\
    \cline{2-12}

    & 0.75 & $-0.09 $ & $ 0.73 $ & $ 0.03 $ & $ 0.38 $ & $ -0.12 $ & $ 2.64 $ & $ -0.22 $ & $ 2.87 $ & $ 0.5 $ & $ 0.0 $ \\
    \cline{2-12}

    \hline\hline

    \multirow{3}{*}{\shortstack[l]{Match distance\\ $d_{\mathrm{MTC}}$ in \SI{}{\metre}}}
    
    & 1 & $-0.06 $ & $ \boldsymbol{0.32} $ & $ 0.03 $ & $ 0.24 $ & $ -0.16 $ & $ 2.63 $ & $  \boldsymbol{0.02} $ & $  \boldsymbol{1.39} $ & $ 0.13 $ & $ -0.74 $\\
    \cline{2-12}

    & 2 & $-0.04 $ & $ 0.50 $ & $ 0.03 $ & $ 0.30 $ & $ -0.11 $ & $ 2.56 $ & $ -0.04 $ & $ 2.02 $ & $ 0.23 $ & $ -0.54 $ \\
    \cline{2-12}

    & 7 & $-0.08 $ & $ 0.80 $ & $ 0.03 $ & $ 0.40 $ & $ -0.11 $ & $\boldsymbol{2.55} $ & $ -0.29 $ & $ 3.60 $ & $ 0.53 $ & $ 0.06 $\\
    \cline{2-12}

    \hline\hline
    
    \multirow{3}{*}{\shortstack[l]{Match counter\\threshold $t_{\mathrm{MTC}}$}}
    
    & 6 & $-0.07 $ & $ 0.71 $ & $ 0.03 $ & $ 0.35 $ & $ -0.11 $ & $ 2.56 $ & $ -0.17 $ & $ 2.81 $ & $ 0.38 $ & $ -0.24 $ \\
    \cline{2-12}

    & 12 & $-0.08 $ & $ 0.73 $ & $ 0.03 $ & $ 0.37 $ & $ -0.09 $ & $ 2.59 $ & $ -0.21 $ & $ 2.82 $ & $ 0.46 $ & $ -0.08 $ \\
    \cline{2-12}

    & 37 & $-0.08 $ & $ 0.74 $ & $ 0.03 $ & $ 0.38 $ & $ -0.11 $ & $ 2.58 $ & $ -0.23 $ & $ 2.96 $ & $ 0.5 $ & $ 0.0 $ \\
    \cline{2-12}
    
    \hline\hline
    
    \multirow{3}{*}{\shortstack[l]{Filter frequency\\ $f_{\mathrm{EKF}}$ in \SI{}{\hertz}}}
    
    & 10 & $-0.06 $ & $ 1.51 $ & $ 0.02 $ & $ 0.33 $ & $ -0.42 $ & $ 2.62 $ & $ -0.22 $ & $ 3.34 $ & $ 0.31 $ & $ -0.38 $\\
    \cline{2-12}
    & 50 & $-0.06 $ & $ 0.74 $ & $ 0.03 $ & $ 0.36 $ & $ -0.15 $ & $\boldsymbol{2.55} $ & $ -0.20 $ & $ 2.98 $ & $ 0.53 $ & $ 0.06 $ \\
    \cline{2-12}
    & 200 & $ -0.08 $ & $ 0.81 $ & $ 0.03 $ & $ 0.41 $ & $ -0.08 $ & $ 2.64 $ & $ -0.24 $ & $ 2.74 $ & $ 0.52 $ & $ 0.03 $ \\
    
    \hline\hline

    \multirow{3}{*}{\shortstack[l]{Merge distance \\ $d_{\mathrm{MRG}}$ in \SI{}{\metre}}}
    
    & 1.7 & $-0.08 $ & $ 0.69 $ & $ 0.03 $ & $ 0.37 $ & $ -0.12 $ & $ 2.60 $ & $ -0.14 $ & $ 2.73 $ & $ 0.36 $ & $ -0.28 $ \\
    \cline{2-12}
    & 3.4 & $-0.09 $ & $ 0.72 $ & $ 0.03 $ & $ 0.37 $ & $ -0.12 $ & $ 2.58 $ & $ -0.20 $ & $ 2.93 $ & $ 0.46 $ & $ -0.08 $ \\
    \cline{2-12}
    & 7.2 & $-0.06 $ & $ 0.71 $ & $ 0.03 $ & $ 0.38 $ & $ -0.10 $ & $ 2.60 $ & $ -0.19 $ & $ 2.61 $ & $ 0.53 $ & $ 0.06 $ \\
    
    \hline\hline

    \multirow{2}{*}{\shortstack[l]{Sensor modality}}
    
    & C & $  \boldsymbol{0.00} $ & $ 0.61 $ & $ \boldsymbol{0.00} $ & $ \boldsymbol{0.22} $ & $ \boldsymbol{-0.07}  $ & $ \boldsymbol{2.55} $ & $ - $ & $ - $ & $  \boldsymbol{0.59} $ & $  \boldsymbol{0.17} $ \\
    \cline{2-12}
    
    & R & $-0.20 $ & $ 1.18 $ & $ -0.03 $ & $ 0.97 $ & $ -0.27 $ & $ 3.57 $ & $ -0.08 $ & $ 2.30 $ & $ 0.32 $ & $ -0.36 $\\
    \cline{2-12}

    \hline
    
    \end{tabular}
    \label{tab:sensitivity}
\end{table*}

We focus the sensitivity analysis on the following parameters: the frequency of the ROS2 node, the distance to the walls of the out-of-bounds filter, the matching distance between estimation and measurement, the upper threshold of the match counter for removal of old unmatched objects, the frequency of the EKF, and the merge distance between neighboring detections so that they can be merged. Additionally, we analyze the tracking behavior for each modality separately to evaluate the respective influence. Table \ref{tab:sensitivity} contains the results.

The variation of the ROS2 node frequency shows that above a frequency of \SI{50}{\hertz}, no improvement of the measurement residuals is reached. The main advantages of a higher frequency are lower perception delays and a lower number of missed published object lists by the respective detection pipelines due to the non-synchronized software execution. In the given dataset, the number of subscribed RADAR detections increases by \qty{2.7}{\percent} if the module is executed at \SI{100}{\hertz}. The number of subscribed LiDAR clustering detections remains the same due to the lower frequency of this algorithm.

The variation of the out-of-bounds filter distance mainly influences the FP rate. The precision deteriorates by \qty{15}{\percent} if the out-of-bounds filter is set to \SI{0}{\metre}. Therefore, the majority of FPs are detected close to the track boundaries.

A low matching distance improves the residuals of the longitudinal position and the velocity. This correlation is reasonable because only detections close to the estimation are considered. However, there is a trade-off between the FP rate and the measurement residuals, because the precision gets significantly worse with a lower matching distance and better with a higher distance.

The upper threshold of the match counter, which indicates the limit for the number of non-detections before an object is removed from the object storage, is set to 25 by default, which correlates to approximately \SI{2}{\second} without successful rematching of the object. The reduction to 12 and 6 shows an increase in the FP-rate. A match counter threshold set to 37 does not affect the FP-rate. However, the raw log data reveals that, as expected, the objects are kept longer in storage before they are removed. This leads to the conclusion that FPs mainly result from short observations with match counters below 25.

Independent of the node frequency, the prediction step of the EKF's kinematic model is conducted at its own frequency. By default, this filter frequency is \SI{100}{\hertz}. If we double the frequency, the measurement residuals decrease slightly. This can be explained by the influence of the yaw angle on the position estimate. The non-linearity leads to different position updates. In particular, the negative influence of the non-measured and thus highly uncertain yaw rate on the yaw estimation of the object increases with increasing frequency (see eq. \ref{eq:ctrv}). In contrast, a filter frequency lower than \SI{50}{\hertz} negatively impacts the measurement residuals because the object's dynamic is insufficiently modeled.

The merge distance between neighboring detected objects which is applied to merge them into one is implemented to reduce the number of FPs around one TP object. The lower precision confirms the effect that more objects are initialized close to each other if the distance is reduced. Increasing the merge distance increases the precision. However, even though a high merge distance is acceptable in the given ODD, it is not applicable to ODDs with dense traffic, such as urban environments. The residuals of the measured features do not vary significantly with the merging distance.

The results of the single sensor modality tracking reveal the influences of the respective single sensors. The LiDAR clustering (C) algorithm is highly accurate in measuring the object's position, which can be seen by the decrease in the standard deviation and shift towards a zero-mean of the longitudinal and lateral residuals. The more precise position measurement also improves the tracking of the object's heading. In addition, the LiDAR algorithm has a lower number of false positives. So an increase in the tracking precision can be observed. 
Compared to the parameterization of the measurement uncertainty (see Table II), the longitudinal standard deviation is too low, while the lateral standard deviation is slightly too high.
The standard deviation of the yaw angle is significantly lower than that assumed by the parameterization. The RADAR detection's (R) biggest benefit is the ability to measure the object's speed. The single modality run reveals a reduction in the velocity residual even though the position and yaw residuals become significantly worse, which degrades the velocity prior and in doing so also influences the velocity residual. The precision decreases by \qty{36}{\percent} due to the high false positive rate of the RADAR. In comparison to the filter parameterization (Table \ref{tab:parameter_tracking}), the longitudinal, the lateral, and the yaw standard deviation are modeled with higher uncertainty than resulted from the data evaluation. 
In contrast, the assumed measurement uncertainty of the velocity is too small by an order of magnitude when compared with real-world results.

The single sensor modality tracking also reveals the influence of sensor noise. The RADAR has more sensor noise than the LiDAR, which can be seen by the higher deviations. In lateral direction, the algorithm is robust against the noise, because the mean of the residuals stays close to zero. In contrast, the sensitivity of the sensor noise in longitudinal direction and on the yaw angle is high as the shift of the mean residuals indicates. However, the noise in form of FPs has the biggest influence, which is shown by the decrease in the precision if only the RADAR is used.

In conclusion, the analysis shows that the precision is mainly influenced by the matching distance, the match counter threshold, and the merge distance.
The respective parameters are set to optimize the trade-offs needed to achieve stable matching of new objects while preventing the holding of objects without detections for too long and minimizing the influence of multiple detections of one object.
The respective parameters are set to balance the trade-off to match stable on new objects, but to minimize the influence of multiple detections for one object and to hold objects without detections for too long. The filter frequency directly manipulates the forward propagation of the kinematic state, which degrades the precision and residuals if a lower frequency is used. The single sensor modality simulations demonstrate their respective influences. In summary, it is apparent that the measurements from the LiDAR sensor are essential for the algorithm to track objects accurately. However, due to the ability to measure the objects' speed, the RADAR is an important associate sensor, and its position measurement uncertainty can be optimized to reduce the induced positional noise.

%% file: source/05_conclusion_contrib.tex
\section{Conclusion} \label{sec:conclusion}
A comprehensive multi-modal late fusion and tracking algorithm is presented and evaluated on high-speed real-world data from the AC@CES 2022.
The modular late fusion is able to consider heterogeneous, multi-modal detection input for an optimized, unique object list. In the shown case, LiDAR and RADAR detections with different update rates, measured features, and detection sensitivities are fused.
Additionally, a novel delay compensation method is presented, which compensates for perception latency by backward-forward integration of the object storage. With this concept, the implemented delay compensation enables the synchronization of the tracked object states with the ego time stamp while still considering delayed perception inputs of up to \SI{325}{\milli\second}. It is one of the core features for enabling high-speed multi-vehicle racing.

Additionally, reliable tracking of detected objects with low-biased measurement residuals and solid precision in real-world data evaluation is demonstrated.
It is the first published real-world application of a multi-modal late fusion and tracking algorithm in multi-object racing at high speeds. The application at the AC@CES proves its applicability and robustness. Besides that, it should be mentioned that the development of the method was realized without labeled or ground truth data compared to other approaches based on neural networks, which is a major factor regarding the application effort for further use cases.

One further research direction is the improvement of the transient behavior until an object is tracked stably. The increased tracking residuals at short observation times lead to short object tracks, i.e., objects are discarded multiple times before they are consistently tracked. The optimization of the state initiation and adaptive parameterization could be future steps to improve the transient phase. The adaptive parameterization could be in relation to the object's observation time, the relative location of the object in front or behind the ego object, or distance-based as the accuracy of the detection algorithms increases with decreased ego-object distance.

As shown in the results, an alternative kinematic model, such as the CTRA-model could improve the state estimation accuracy, which is another possible future step. However, if not measured, the oscillations of the acceleration value could destabilize the state estimation.

Another important factor is the matching method, which is essential to reliably track an object stable. The simple but robust implementation of the Euclidean distance-based matching logic could be improved by more comprehensive metrics such as the Mahalanobis distance \cite{DEMAESSCHALCK20001}. However, the robustness of real-world applications remains to be investigated.

The entire code and recorded real-world data are publicly available. The source code additionally comprises a ROS2 launch configuration and a Docker build file.

\section*{Contributions}
As the first author, Phillip Karle initiated the idea of this paper and contributed essentially to its conception, implementation, and content. Felix Fent, Sebastian Huch, and Florian Sauerbeck contributed to the conception of this research, the experimental data generation and the revision of the research article. Markus Lienkamp made an essential contribution to the conception of the research project. He revised the paper critically for important intellectual content. He gave final approval of the version to be published and agreed with all aspects of the work. As a guarantor, he accepts responsibility for the overall integrity of the paper.

%% file: source/98_appendix.tex
{

\appendix[Parameterization]

\begin{table}[htbp]
\caption{Parameterization of the proposed approach.}
\begin{center}

\begin{tabular}{|c|c|c|} \hline
\textbf{Parameter} & \textbf{Value} & \textbf{Description} \\ \hline

$f_{\mathrm{node}}$ & \SI{50}{\hertz} &  ROS2-node frequency \\ \hline

$\boldsymbol{z}_{\mathrm{C}}$ & $\left( x, y, \psi \right)$ & Measured values C  \\ \hline
$x_{\mathrm{std, C}}$&\SI{0.3}{\metre} & Lon STD C \\ \hline
$y_{\mathrm{std, C}}$&\SI{0.3}{\metre} & Lat STD C \\ \hline
$\psi_{\mathrm{std, C}}$&\SI{20.0}{\degree} & Yaw STD C \\ \hline

$\boldsymbol{z}_{\mathrm{R}}$ & $\left( x, y, \psi, v \right)$ & Measured values R  \\
\hline
$x_{\mathrm{std, R}}$&\SI{3.0}{\metre} & Lon STD R \\
\hline
$y_{\mathrm{std, R}}$&\SI{3.0}{\metre} & Lat STD R \\
\hline
$\psi_{\mathrm{std, R}}$&\SI{20.0}{\degree} & Yaw STD R \\
\hline
$v_{\mathrm{std, R}}$&\SI{0.2}{\metre\per\second} & Velocity STD R \\ \hline

$f_{\mathrm{EKF}}$ & $ \SI{100}{\hertz}$ & Filter frequency \\ \hline

$x_{\mathrm{std, P}}$&\SI{0.01}{\metre} & Lon initial state STD \\ \hline
$y_{\mathrm{std, P}}$&\SI{0.01}{\metre} & Lat initial state STD  \\ \hline
$\psi_{\mathrm{std, P}}$&\SI{17.2}{\degree} & Yaw initial state STD \\ \hline
$v_{\mathrm{std, P}}$&\SI{4.0}{\metre\per\second} & Velocity initial state STD  \\ \hline
$\dot{\psi}_{\mathrm{std, P}}$&\SI{17.2}{\degree\per\second} & Yaw-Rate initial state STD \\ \hline
$ k_v$ & $ 0.8 $ & Ego speed factor \\ \hline

$\dot{x}_{\mathrm{std, P}}$&\SI{0.01}{\metre} & Lon initial state STD \\ \hline
$\dot{y}_{\mathrm{std, P}}$&\SI{0.01}{\metre} & Lat initial state STD  \\ \hline
$\dot{\psi}_{\mathrm{std, P}}$&\SI{17.2}{\degree} & Yaw initial state STD \\ \hline
$\dot{v}_{\mathrm{std, P}}$&\SI{4.0}{\metre\per\second} & Velocity initial state STD  \\ \hline
$\ddot{\psi}_{\mathrm{std, P}}$&\SI{17.2}{\degree\per\second} & Yaw-Rate initial state STD \\ \hline

$d_{\mathrm{MRG}}$&\SI{5.1}{\metre} & MRG distance \\ \hline

$d_{\mathrm{OBF, out}}$&\SI{0.3}{\metre} & Bounds buffer OBF, outside \\ \hline
$d_{\mathrm{OBF, in}}$&\SI{0.3}{\metre} & Bounds buffer, OBF, inside \\ \hline

$d_{\mathrm{MTC}}$&\SI{0.3}{\metre} & MTC distance \\ \hline
$t_{\mathrm{MTC}}$&25 & MTC threshold \\ \hline

\end{tabular}

\vspace{10.0pt}

\begin{tabular}{|c|c|} \hline
\textbf{Abbreviation} & \textbf{Description} \\ \hline
C & LiDAR Clustering \\ \hline
LAT & Lateral \\  \hline
LON & Longitudinal \\  \hline
MTC & Match \\ \hline
MRG & Merge \\ \hline
OBF & Out-of-Bounds Filter \\  \hline
STD & Standard deviation \\  \hline
R & RADAR \\  \hline
\end{tabular}

\label{tab:parameter_tracking}
\end{center}
\end{table}

 }

%% file: source/99_biography.tex
\newpage
\section{Biography Section}
\vspace{-30pt}
\begin{IEEEbiography}
[
{\includegraphics[width=1in,height=1.25in,clip,keepaspectratio]{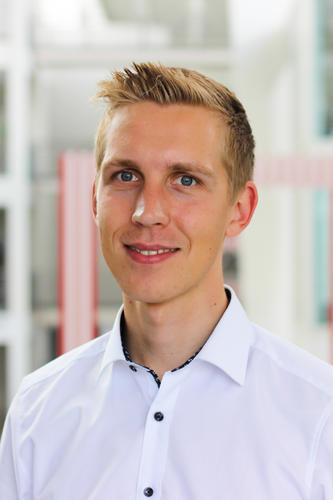}}
]{Phillip Karle}
received the B.Sc. and M.Sc. degrees from the Technical University of Munich (TUM), Munich, Germany, in 2017 and 2019, respectively, where he is currently pursuing a Ph.D. degree in
mechanical engineering with the Institute of Automotive Technology. His research interests include multi-object tracking, scenario understanding, motion prediction, and related applications for autonomous driving with a focus on real-world applications.
\end{IEEEbiography}
\vspace{-30pt}
\begin{IEEEbiography}
[
{\includegraphics[width=1in,height=1.25in,clip,keepaspectratio]{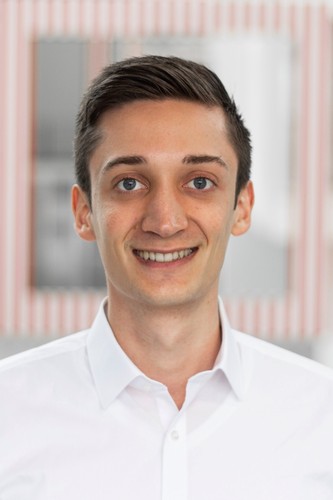}}
]{Felix Fent}
received the B.Sc. and M.Sc. degrees from the Technical University of Munich (TUM), Munich, Germany, in 2018 and 2020, respectively, where he is currently pursuing a Ph.D. degree in
mechanical engineering with the Institute of Automotive Technology. His research interests include radar-based perception, sensor fusion and multi-modal object detection approaches with a focus on real-world applications.
\end{IEEEbiography}
\vspace{-30pt}
\begin{IEEEbiography}
[
{\includegraphics[width=1in,height=1.25in,clip,keepaspectratio]{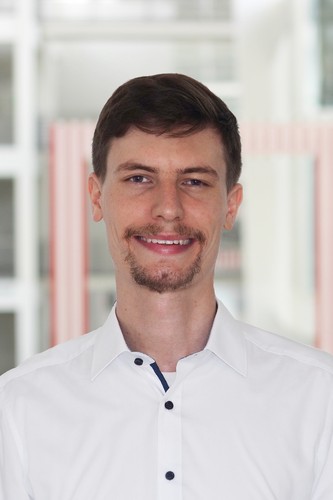}}]{Sebastian Huch}
received his B.E. degree from the Baden-Wuerttemberg Cooperative State University (DHBW) Stuttgart, Germany, in 2016 and his  M.Sc. degree from the Technical University of Darmstadt, Germany, in 2018. He is currently pursuing his Ph.D. degree in mechanical engineering at the Institute of Automotive Technology at the Technical University of Munich (TUM), Germany. His research interests include LiDAR simulation, LiDAR perception, and LiDAR domain adaptation for autonomous driving.
\end{IEEEbiography}
\vspace{-30pt}
\begin{IEEEbiography}
[
{\includegraphics[width=1in,height=1.25in,clip,keepaspectratio]{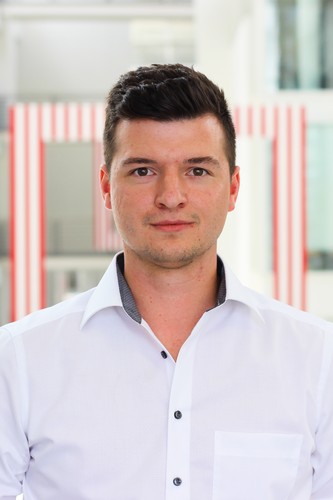}}]{Florian Sauerbeck}
received a B.Sc. from the University of Erlangen–Nuremberg in 2016 and a M.Sc. in Electrical Engineering from the Technical University of Munich (TUM), Munich, Germany, in 2020. 
Currently, he is pursuing a Ph.D. degree in mechanical engineering with the Institute of Automotive Technology at TUM. His research interests include 3D perception, localization, and mapping with a focus on real-world applications.
\end{IEEEbiography}
\vspace{-30pt}
\begin{IEEEbiography}
[
{\includegraphics[width=1in,height=1.25in,clip,keepaspectratio]{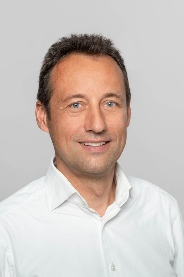}}]{Markus Lienkamp}
researches in the area of autonomous vehicle with the objective to create an open-source software platform. He is professor of the Institute of Automotive Technology at the Technical University of Munich (TUM) since November 2009. After studying mechanical engineering at TU Darmstadt and Cornell University, Prof. Lienkamp obtained his doctorate from TU Darmstadt (1995). 
Afterward, he joined an international trainee program at Volkswagen and worked in a joint venture between Ford and Volkswagen in Portugal, led the brake testing department of the VW commercial vehicles, and was later appointed head of the Electronics and Vehicle research department.
\end{IEEEbiography}